\documentclass[runningheads]{llncs}
\usepackage{booktabs} 
\usepackage{amssymb}   
\usepackage{multirow}
\usepackage{esvect}  
\usepackage{bm}       
\usepackage{amsmath} 
\usepackage[T1]{fontenc}
\usepackage{url}
\usepackage{graphicx}
\begin{document}
\title{YOLO-FireAD: Efficient Fire Detection via Attention-Guided Inverted Residual Learning and Dual-Pooling Feature Preservation}

%
%
\authorrunning{W. Pan et al.} 

\titlerunning{YOLO-FireAD}

\author{
  Weichao Pan\textsuperscript{1}, 
  Bohan Xu\textsuperscript{2},
  Xu Wang\textsuperscript{1},
  Chengze Lv\textsuperscript{1},\\
  Shuoyang Wang\textsuperscript{1},
  Zhenke Duan\textsuperscript{3}\thanks{Corresponding author:Zhenke Duan \texttt{duanzhenke@sscapewh.com}},
  Zhen Tian\textsuperscript{4}, 
}
 
\institute{
  School of Computer Science and Technology, Shandong Jianzhu University\and
  International School of Information Science\&Engineering, Dalian University of Technology\and 
  Wuhan SecureScape Technology Co., Ltd\and
  University of Glasgow, Glasgow, UK
}


\maketitle              
\begin{abstract}
Fire detection in dynamic environments faces continuous challenges, including the interference of illumination changes, many false detections or missed detections, and it is difficult to achieve both efficiency and accuracy. To address the problem of feature extraction limitation and information loss in the existing YOLO-based models, this study propose \textbf{Y}ou \textbf{O}nly \textbf{L}ook \textbf{O}nce for \textbf{Fire} Detection with \textbf{A}ttention-guided Inverted Residual and \textbf{D}ual-pooling Downscale Fusion (\textbf{YOLO-FireAD}) with two core innovations: (1) \textbf{A}ttention-guided \textbf{I}nverted \textbf{R}esidual Block (\textbf{AIR}) integrates hybrid channel-spatial attention with inverted residuals to adaptively enhance fire features and suppress environmental noise; (2) \textbf{D}ual \textbf{P}ool \textbf{D}ownscale \textbf{F}usion Block (\textbf{DPDF}) preserves multi-scale fire patterns through learnable fusion of max-average pooling outputs, mitigating small-fire detection failures. Extensive evaluation on two public datasets shows the efficient performance of our model. Our proposed model keeps the sum amount of parameters (1.45M, 51.8\% lower than YOLOv8n) (4.6G, 43.2\% lower than YOLOv8n), and mAP75 is higher than the mainstream real-time object detection models YOLOv8n, YOL-Ov9t, YOLOv10n, YOLO11n, YOLOv12n and other YOLOv8 variants 1.3-5.5\%. For more details, please visit our repository: https://github.com/JEFfersusu/YOLO-FireAD
\end{abstract}

\section{Introduction}
Fire disasters pose significant threats to human safety, ecological systems, and critical infrastructure worldwide. With increasing urbanization and climate change impacts, developing rapid and reliable fire detection systems has become a crucial research frontier in computer vision and public safety\cite{bib1}. 

Traditional sensor-based approaches often suffer from limited coverage and environmental sensitivity, while emerging vision-based methods leveraging deep learning have demonstrated superior adaptability in complex scenarios\cite{bib2}. However, existing convolutional neural network (CNN) architectures for fire detection face three fundamental challenges in practical deployments: 1) Severe performance degradation under illumination variations and fire-like object interference, 2) The trade-off between effectiveness and efficiency, and 3) Information loss during feature downsampling particularly detrimental to small fire detection.

Current YOLO-based solutions\cite{bib3}, while achieving notable real-time performance, exhibit critical limitations. The standard residual blocks in lightweight networks inadequately capture long-range contextual dependencies crucial for distinguishing fire features from complex backgrounds. Furthermore, conventional pooling operations in feature pyramid networks tend to discard spatial details during downsampling, leading to suboptimal performance in detecting early-stage fires with limited visual signatures. Recent attempts incorporating attention mechanisms or multi-scale fusion either introduce prohibitive computational overhead or fail to maintain feature integrity across scales\cite{bib4,bib5,bib6,bib7}.

To address these challenges, this study propose YOLO-FireAD, a novel fire detection framework that synergistically integrates two specialized modules. Specifically, the contributions of this study are as follows.

1. This study proposes the \textbf{A}ttention-guided \textbf{I}nverted \textbf{R}esidual Block (\textbf{AIR}), which synergistically embeds spatial-channel hybrid attention into inverted residual operations.  This design achieves adaptive fire feature amplification through gated position-aware convolutions and channel reweighting, reducing false alarms from fire-like interference while maintaining computational efficiency (39\% parameter reduction vs. baseline).

2. This study proposes the \textbf{D}ual \textbf{P}ool \textbf{D}ownscale \textbf{F}usion Block (\textbf{DPDF}) addresses small-fire detection failures through parallel max-average pooling fusion with learnable coefficients.   This strategy preserves both flame edge details and smoke continuity during downsampling, improving mAP50 by 1.7\% compared to standard YOLOv8n while reducing GFLOPs by 15\%.

3. This study uses common evaluation metrics combined with cross-dataset testing (fire\_detection dataset and fire dataset) for validation. Extensive experiments show that the model proposed in this study, YOLO-FireAD model, has a good precisionefficiency balance, achieving 34.6\% mAP50-95 (1.8 \% higher than YOLOv8) with 51.8\% parameter reduction.

The remainder of this paper is organized as follows: Section 2 reviews related work in fire detection and lightweight CNN architectures. Section 3 details the technical implementation of AIR and DPDF modules. Section 4 presents experimental setup and comparative analysis. Section 5 discusses practical implications and limitations, followed by concluding remarks in Section 6.

\section{Methods}

In this section, we offer a comprehensive description of each module within the network model, clarifying their specific functions. We also provide an overall explanation of the model, detailing the involved components and structures, including the AIR block and DPDF block.

\begin{figure}
    \centering
    \includegraphics[width=0.83\linewidth]{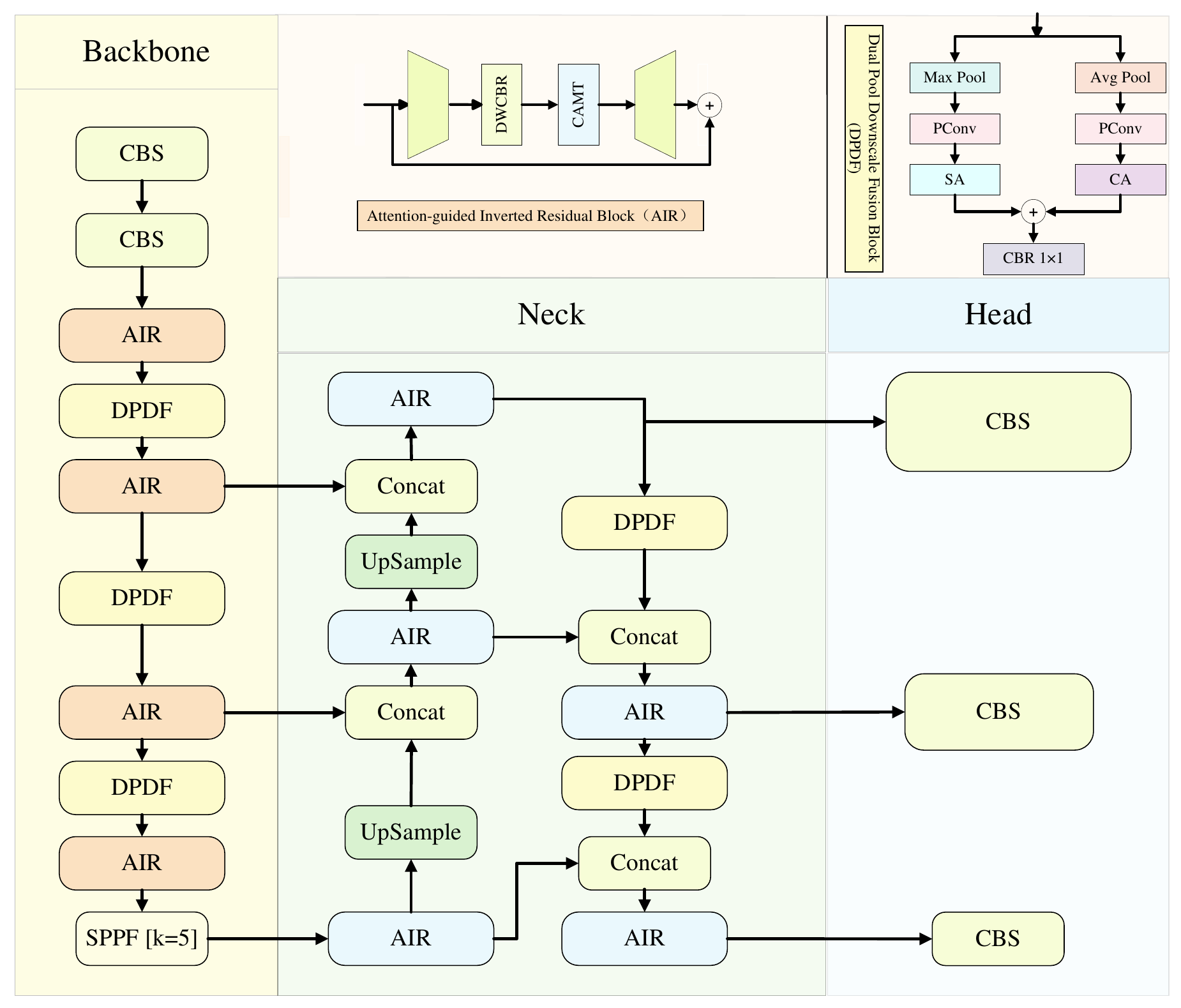}
    \caption{Overview of the YOLO-FireAD model architecture.}
    \label{fig1}
\end{figure}

\subsection{Overview}

As shown in Figure 1, YOLO-FireAD adopts a "Backbone-Neck-Head" three-tier architecture, which is optimised for early fire detection and high efficiency. Its core lies in combining the attention-guided feature enhancement of the AIR module and the multi-granularity feature retention mechanism of the DPDF module throughout the entire network, which improves the mAP50-95 of fire detection to 34.6\% (1.8\% higher than that of YOLOv8\cite{bib8}) while maintaining high efficiency (53\% lower than the number of YOLOv8 parameters).

In the Backbone stage, based on the multi-stage feature extraction design, it consists of CBS base module, AIR attention module and DPDF downsampling module stacked alternately: the CBS module extracts the local texture features of the flame (such as edge flickering, smoke diffusion pattern) through 3×3 convolution; the AIR module embeds the attention mechanism, which enhances the response of the flame core region and suppresses the fire-like interference; and the DPDF module preserves the flame highlight details (maximum pooling) and smoke continuity (average pooling) simultaneously during resolution downscaling through a dual-pooling fusion strategy.

A bi-directional feature pyramid structure is used in the neck stage to fuse deep semantics with shallow details through UpSample and Concat operations.The DPDF module performs resolution alignment and noise filtering at the feature fusion nodes to remove feature blurring due to smoke occlusion.The AIR module inserts cross-layer connectors to dynamically calibrate the attentional distributions of features at different scales.

The detection head, while maintaining multi-scale detection capability, achieves channel compression by stacking CBS modules and finally enters the loss function CIoU loss\cite{bib9} as well as cross-entropy loss for localisation and classification.

\subsection{Attention-guided Inverted Residual Block (AIR)}

Aiming at the problem of confusion between fire-like interferences (such as light reflections) and real flames in flame detection, this study proposes the AIR module, which inherits and improves the classical inverted residual structure, and embeds the lightweight channel-spatial hybrid attention mechanism into the feature transformation process to achieve efficient feature discrimination through the three-stage design of feature decoupling-attention enhancement-lightweight compression. The module structure is shown in Figure 2.

\begin{figure}
    \centering
    \includegraphics[width=0.5\linewidth]{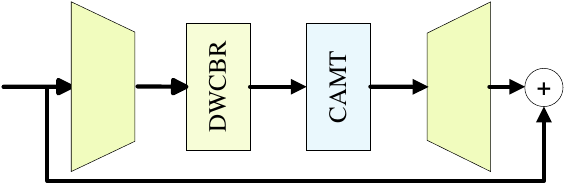}
    \caption{Overview of the Attention-guided Inverted Residual Block structure.}
    \label{fig2}
\end{figure}

For the input features $X_{in} \in \mathbb{R}^{C \times H \times W}$, firstly, the channel dimension reduction is achieved by reducing the convolution to reduce the amount of calculation:
\begin{equation}
    X_{reduce} = \text{ReLU}\left( \text{BN}\left( \text{Conv}_{1 \times 1}(X_{in}) \right) \right)
\end{equation}

The reduction rate $r = 0.25$ here is experimentally optimised to make a balance between reducing information loss and computational complexity. The deep feature extraction stage uses a deep separable convolution\cite{bib10} with more efficient parameterisation:
\begin{equation}
    X_{dw} = \text{ReLU}\left( \text{BN}\left( \text{DepthwiseConv}_{3 \times 3}(X_{reduce}) \right) \right)
\end{equation}

This operation extracts local features via a spatial convolution kernel $k \times k$ while maintaining channel independence, reducing the computational complexity from $O(k^2 C_{exp}^2)$ to $O(k^2 C_{exp})$.

\begin{figure}
    \centering
    \includegraphics[width=0.5\linewidth]{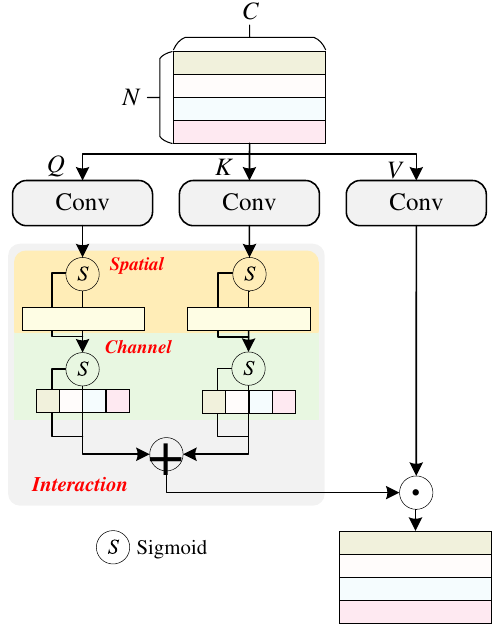}
    \caption{Overview of the convolutional additive self-attention structure.}
    \label{fig3}
\end{figure}

Then $X_{dw} $ undergo convolutional additive self-attention\cite{bib11} through the following process: First, the input features are decomposed into query (Q), key (K), and value (V) triples by 1×1 convolution:
\begin{equation}
    [Q, K, V] = \text{Split}\left(\text{Conv}_{1 \times 1}(X)\right)
\end{equation}

Then the parallel processing strategy of space-channel decoupling is adopted:
\begin{align}
    \hat{Q} &= \mathcal{F}_{channel}\left(\mathcal{F}_{spatial}(Q)\right) \\
    \hat{K} &= \mathcal{F}_{channel}\left(\mathcal{F}_{spatial}(K)\right)
\end{align}
where the spatial operation $\mathcal{F}_{spatial}$ uses gated convolution to enhance position perception:
\begin{equation}
    \mathcal{F}_{spatial}(x) = x \odot \sigma\left(\text{Conv}_{3 \times 3}\left(\text{DepthwiseConv}(x)\right)\right)
\end{equation}

The channel operation $\mathcal{F}_{channel}$ establishes global dependencies through adaptive pooling:
\begin{equation}
    \mathcal{F}_{channel}(x) = x \odot \sigma\left(\text{MLP}\left(\text{GAP}(x)\right)\right)
\end{equation}

Then feature enhancement is achieved by additive fusion and deep convolution:
\begin{equation}
    X_{out} = \text{Dropout}\left(\text{DepthwiseConv}_{3 \times 3}\left((\hat{Q} + \hat{K}) \odot V\right)\right)
\end{equation}

where $\odot$ denotes element-by-element multiplication, the design reduces $O(C^2)$ parameter overhead while maintaining feature interaction capabilities.

Finally, the number of channels is restored to facilitate subsequent feature extraction. The module structure is shown in Figure 3.

\subsection{Dual Pool Downscale Fusion Block (DPDF)}

For the problems of small target leakage and dynamic blurring in flame detection, the DPDF module achieves multi-granularity information fusion through a dual-path feature retention mechanism. As shown in Figure 4, the module synchronously preserves flame edge details (via maximum pooling) and smoke diffusion features (via average pooling) during resolution downsampling.

\begin{figure}
    \centering
    \includegraphics[width=0.3\linewidth]{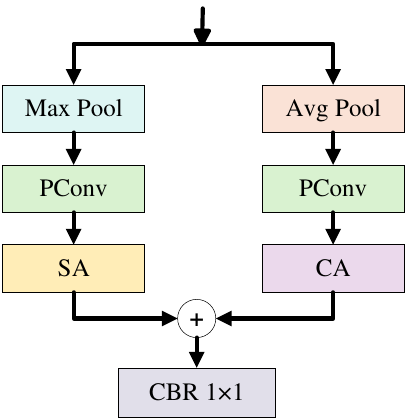}
    \includegraphics[width=0.5\linewidth]{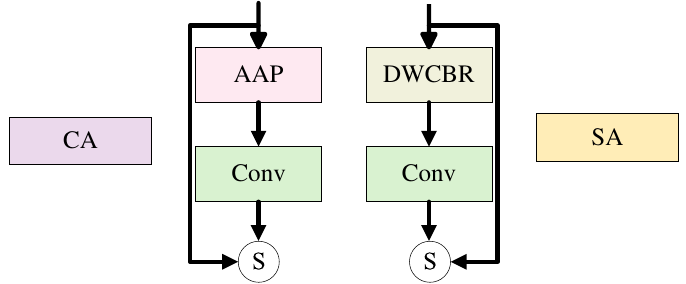}
    \caption{Overview of the Dual Pool Downscale Fusion Block structure.}
    \label{fig:enter-label}
\end{figure}

For dual path feature extraction, two pooling operations are first performed in parallel to obtain complementary features:

\begin{equation}
X_{max} = \text{MaxPool}_{2 \times 2}(X_{in})
\end{equation}

\begin{equation}
X_{avg} = \text{AvgPool}_{2 \times 2}(X_{in})
\end{equation}

The maximum pooling path reinforces the flame spike features (such as flickering highlight areas), and the average pooling path maintains the continuity of smoke diffusion.

Partial convolution (PConv)\cite{bib12} is used to reduce the computational overhead:
\begin{equation}
X_{max}' = \text{PConv}(X_{max}) = \text{DepthwiseConv}_{3 \times 3}(X_{max}^{[:r:]}) \oplus X_{max}^{[r::]}
\end{equation}
where $ r=4 $ is the channel reduction rate, performing deep convolution for only $ 1/4 $ of the channels and passing through the rest. This design reduces FLOPs by 68\% while maintaining 90\% feature accuracy.

Feature calibration is implemented via the SA (Spatial Attention) and CA (Channel Attention) modules:
\begin{equation}
X_{max}'' = \text{CA}(\text{SA}(X_{max}'))
\end{equation}
\begin{equation}
X_{avg}'' = \text{CA}(\text{SA}(X_{avg}'))
\end{equation}

Among them, the SA module enhances the flame edge response using cavity convolution, and the CA module applies channel reweighting based on the flame chromaticity histogram.

Finally, adaptive weighted fusion is achieved through learnable coefficients:
\begin{equation}
X_{fusion} = \alpha \cdot X_{max}'' + (1 - \alpha) \cdot X_{avg}''
\end{equation}

\section{Experimental setup}

In this section, a brief overview of the experimental setup and related resourceswill be given. In the following, this study introduce the experimental dataset, the experimental configuration, and the evaluation metric in turn.

\subsection{Dataset}

The two datasets used in this study, fire\_detection dataset and fire dataset, are both from the Paddle community and contain a large number of multi-scale complex flame fire scenes and smoke.

\subsection{Configuration}

The experimental program was executed on ubuntu22.04 operating system with NVIDIA GeForce RTX 4090 graphics card driver. The deep learning framework was selected as pytorch with 2.1.0, jupyter notebook was used for the compiler, Python 3.10 was used as the specified programming language, and all the algorithms used in the comparative analysis were operationally consistent and ran in the same computational setup. The image size was normalized to 640×640×3, and the batch size was 128, the optimizer was AdamW, the learning rate was set to 0.002, the momentum to 0.9, and the number of training periods was 200.

\subsection{Evaluation metric}

Model performance was evaluated using detection accuracy metrics (precision, recall, f1 score), comprehensive mAP scores (mAP50, mAP75, mAP50-95), and computational efficiency parameters (Params, GFLOPs, Model Size) for holistic assessment\cite{bib13}.

\section{Experimental details and analyses}

In order to validate the superior performance of the YOLO-FireAD detection model proposed in this paper, a series of validations are conducted on the above dataset and evaluated and analyzed using several evaluation metrics mentioned above.

Firstly, the current mainstream object detection models and their variants are introduced, and conducts comparison experiments with the model proposed in this paper to demonstrate the superiority of the proposed model. Then, the results of the model proposed in this paper are evaluated, including comparative experimental analysis, generalization experimental analysis, and ablation study analysis. Finally, the limitations of the model proposed in this paper are analyzed.

\begin{figure}
    \centering
    \includegraphics[width=0.9\linewidth]{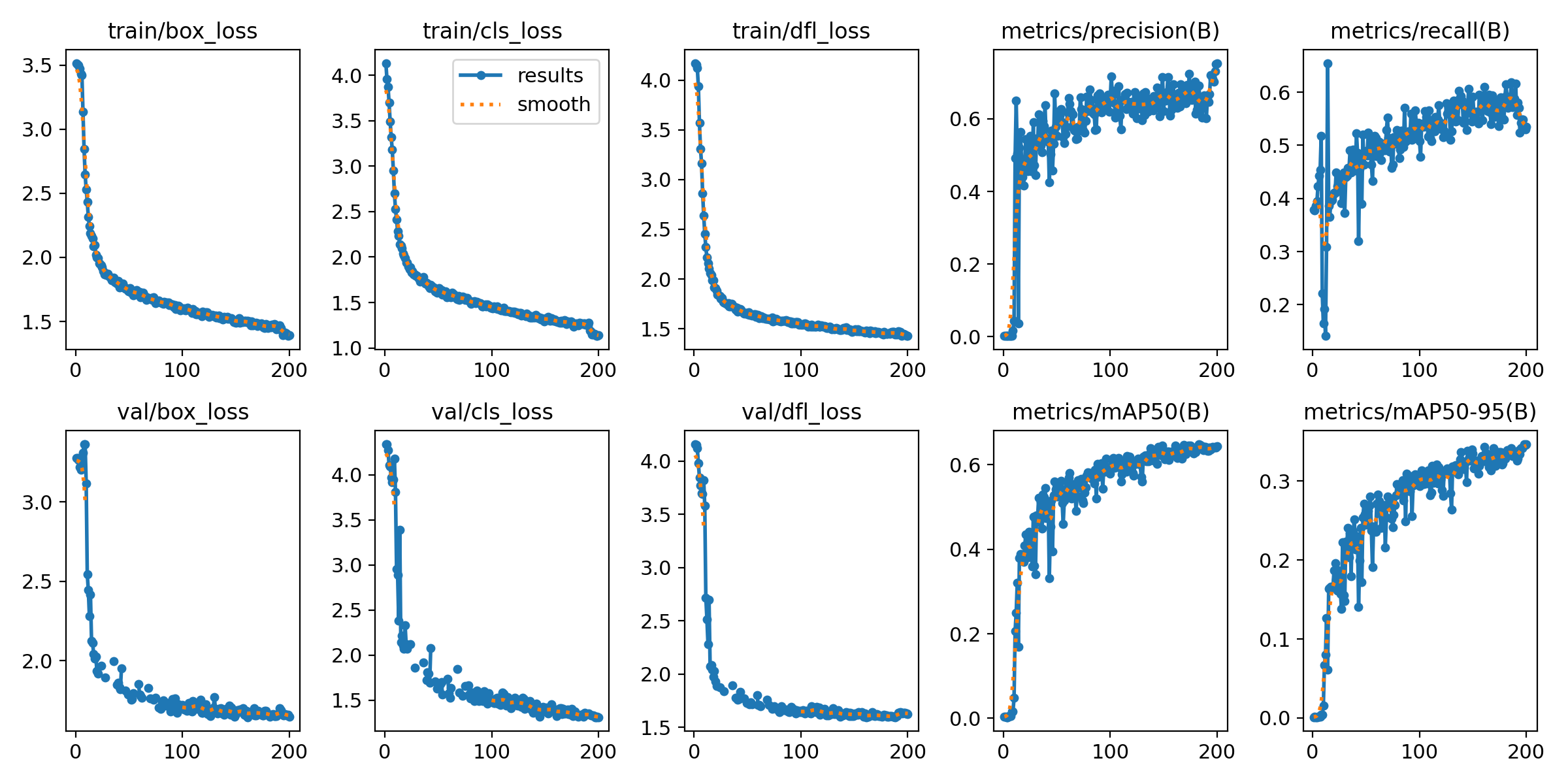}
    \caption{YOLO-FireAD iterations on the fire\_detection dataset.}
    \label{fig5}
\end{figure}

Figure 5 records the changes of the bounding box loss, classification loss, focus loss, model precision, recall, and average precision for different thresholds in each iteration of the model on the training set and the test set. According to the charts, it can be seen that in the early stage of training, the model has high values for each loss and converges quickly, while the model metrics rise rapidly, indicating that the model is learning efficiently at this time. As the number of iterations increases, the rate of change of each curve gradually decreases, and after about 40 iterations, the curves begin to stabilise, and finally about 200 iterations, the curves basically remain stable.

\subsection{Comparative experimental analysis}

\begin{table}[ht]
\centering
\caption{Test results of all models on fire\_detection dataset.}
\begin{tabular}{lccccccccc}
\toprule
\textbf{Model} & \textbf{P} & \textbf{R} & \textbf{F1} & \textbf{mAP50} & \textbf{mAP75} & \textbf{mAP50-95} & \textbf{Params} & \textbf{GFLOPs} & \textbf{Size} \\
\midrule
YOLOv8n & 64.6 & 62.4 & 63.5 & 64.2 & 31.9 & 32.8 & 3.01 & 8.1 & 6.1 \\
YOLOv9t & 66.8 & 60.3 & 63.4 & 64.4 & 31.8 & 33.3 & 1.97 & 7.6 & 4.6 \\
YOLOv10n & 68.2 & 55.5 & 61.2 & 60.0 & 30.0 & 31.5 & 2.70 & 8.2 & 5.6 \\
YOLO11n & 66.5 & 58.7 & 62.4 & 64.3 & 31.5 & 33.6 & 2.58 & 6.3 & 5.4 \\
YOLOv12n & 65.9 & 61.0 & 63.3 & 66.1 & 34.2 & 35.2 & 2.56 & 6.3 & 5.4 \\
YOLOv8n+SE & 70.4 & 59.5 & 64.5 & 63.5 & 31.4 & 33.0 & 3.01 & 8.1 & 6.1 \\
YOLOv8n+HAM & 63.3 & 63.3 & 63.3 & 64.8 & 32.7 & 33.7 & 3.01 & 8.1 & 6.1 \\
YOLOv8n+ASF & 69.1 & 59.8 & 64.1 & 65.1 & 32.8 & 34.2 & 3.05 & 8.6 & 6.2 \\
YOLOv8n+BifPN & 67.6 & 60.9 & 64.1 & 63.4 & 32.3 & 33.6 & 3.02 & 8.1 & 6.1 \\
YOLOv8n+WIoUv3 & 63.9 & 61.5 & 62.7 & 63.4 & 31.7 & 32.7 & 3.01 & 8.1 & 6.1 \\
YOLOv8n+MDPIoU & 65.4 & 61.3 & 63.4 & 63.8 & 32.2 & 33.4 & 3.01 & 8.1 & 6.1 \\
YOLOv8n+PIoUv2 & 66.1 & 60.3 & 63.2 & 63.9 & 33.8 & 33.4 & 3.01 & 8.1 & 6.1 \\
\textbf{Ours} & \textbf{75.3} & \textbf{53.3} & \textbf{64.3} & \textbf{64.9} & \textbf{35.5} & \textbf{34.6} & \textbf{1.45} & \textbf{4.6} & \textbf{3.3} \\
\bottomrule
\end{tabular}
\label{tab1}
\end{table}

In order to verify the performance of YOLO-FireAD, the comparison experiment compares YOLO-FireAD with YOLOv8n\cite{bib8}, YOLOv9t\cite{bib14}, YOLOv10n\cite{bib15}, YOLO11n\cite{bib8}, YOLOv12n, and the improved models of YOLOv8n\cite{bib16,bib17,bib18,bib19,bib20,bib21,bib22} on fire\_detection dataset, and analyse the model performance in terms of P, R, F1score, mAP50, mAP50-75,mAP50-95, Params, FLOPs, Model Size to analyse the model performance, the experimental results are shown in Table 1, and Figure 6 is the visual graph of the experimental data of F1, mAP50 ,Params. It can be seen that the YOLO-FireAD model performs well in all performance indicators, and its P, R, F1, and mAP (mAP50, mAP75, and mAP50-90) are significantly better than the other models. The advantage of P is the most obvious, which is 10.7\%, 9.4\%, 12\%, and 11.4\% higher compared with YOLOv8n, YOLOv12n, YOLOv8n+HAM, and YOLOv8n+WIoUv3, respectively. In addition, YOLO-FireAD's mAP75 performs the best among all models, which indicates that YOLO-FireAD maintains high detection performance at different IoU thresholds, which gives it a strong advantage in complex bad detection tasks. While maintaining high accuracy, the complexity of YOLO-FireAD is also significantly lower than other models, with Params of 1.45M, which is only 48.2\% of YOLOv8n, GFLOPs of 4.6G, and Model Size of 3.3MB, which are the lowest among all the compared models.

\begin{figure}
    \centering
    \includegraphics[width=0.9\linewidth]{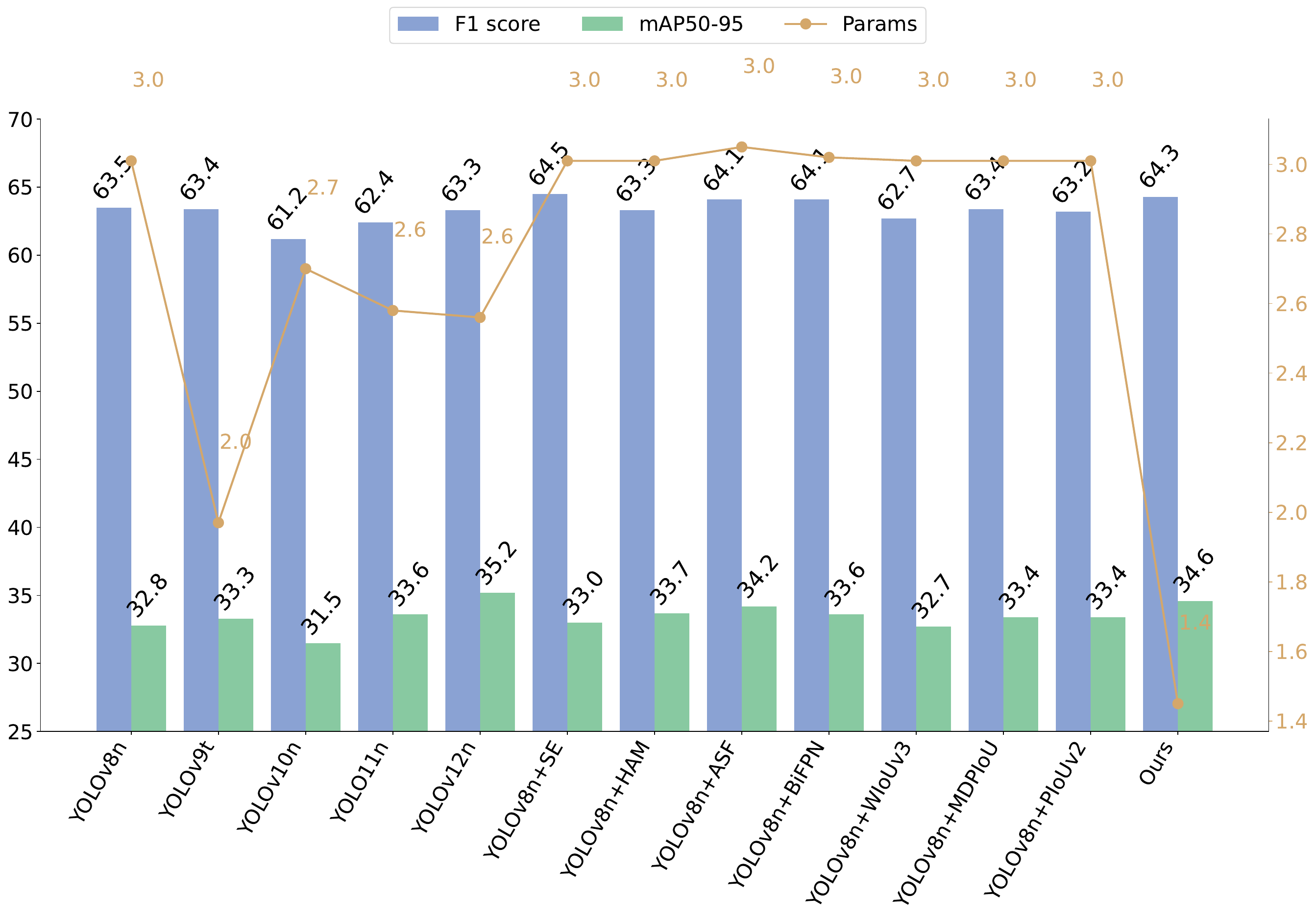}
    \caption{Comparison of model metrics (F1 score, mAP50-95, Params) on fire\_detection dataset.}
    \label{fig6}
\end{figure}

\begin{figure}
    \centering
    \includegraphics[width=0.9\linewidth]{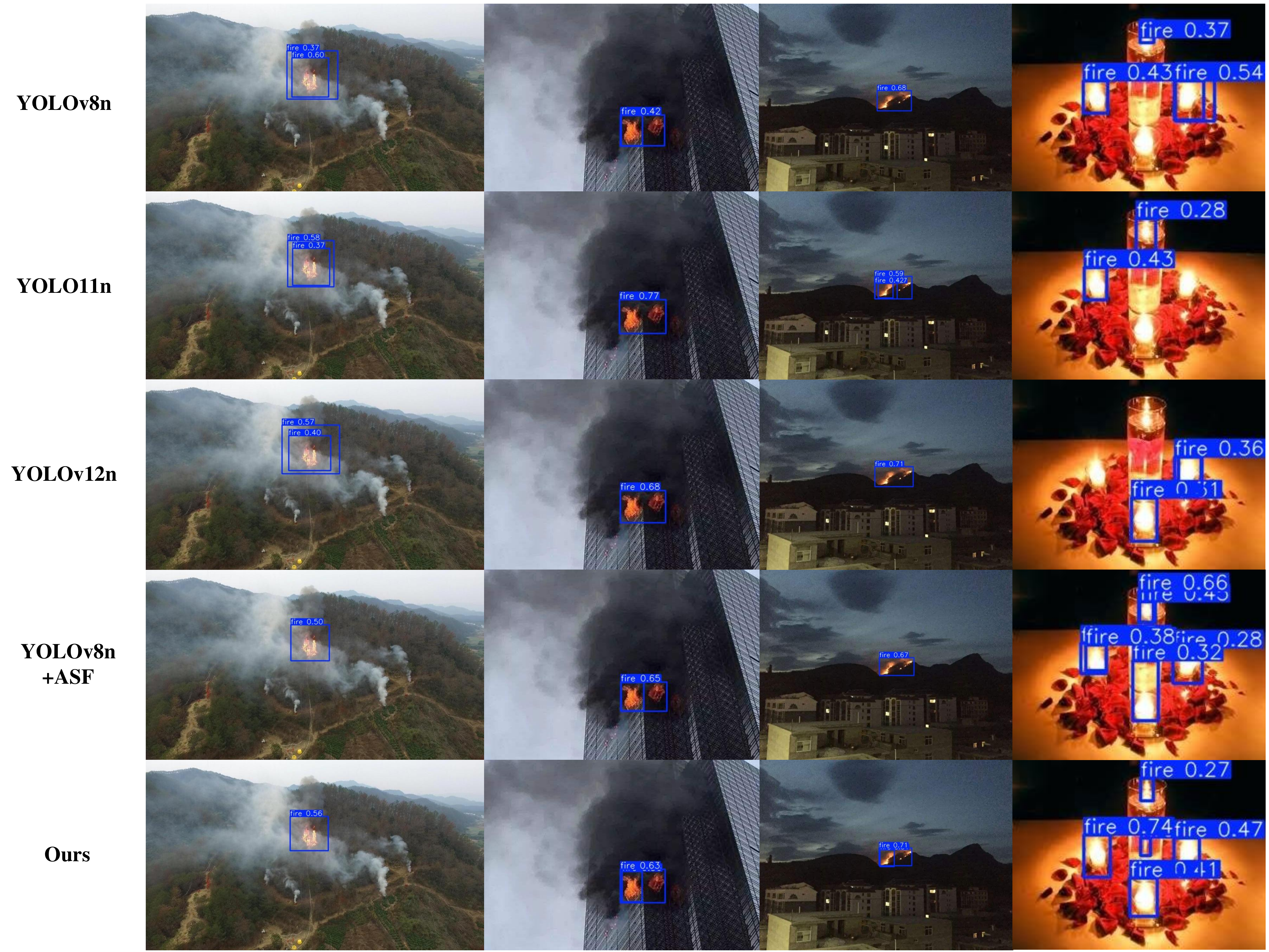}
    \caption{Sample visualisation of several competitive model identification results}
    \label{fig7}
\end{figure}

Figure 7 shows several representative sets of detection results of several mainstream models and YOLO-FireAD in actual situations, in order to clearly see the differences in performance of different models in flame hazard detection. YOLOv8n is able to identify flames in most of the scenarios, but the recognition accuracy is slightly lower, and there is a certain leakage of detection in dense flame scenarios. Compared with YOLOv8n, YOLO11n and YOLOv12n have higher leakage rates in dense flames, which indicates that there are some limitations in their flame recognition in complex scenarios. Compared to the above three models YOLOv8n+ASF has a significantly lower leakage rate when the flame is dense, but there is a certain leakage for small flames, which indicates that it has some limitations when dealing with small-sized objects. Compared with the above models, YOLO-FireAD performs significantly better with the attention mechanism of its AIR module, which can dynamically adjust the weights of the feature maps to make the network more focused on the flame targets, especially in multi-dense flame scenes.The DPDF module synchronously retains the flame highlight details during resolution downscaling through the dual-pooling fusion strategy, avoiding the loss of the small flame features during downsampling, thus achieving a more accurate recognition of small flames.

\begin{figure}
    \centering
    \includegraphics[width=0.9\linewidth]{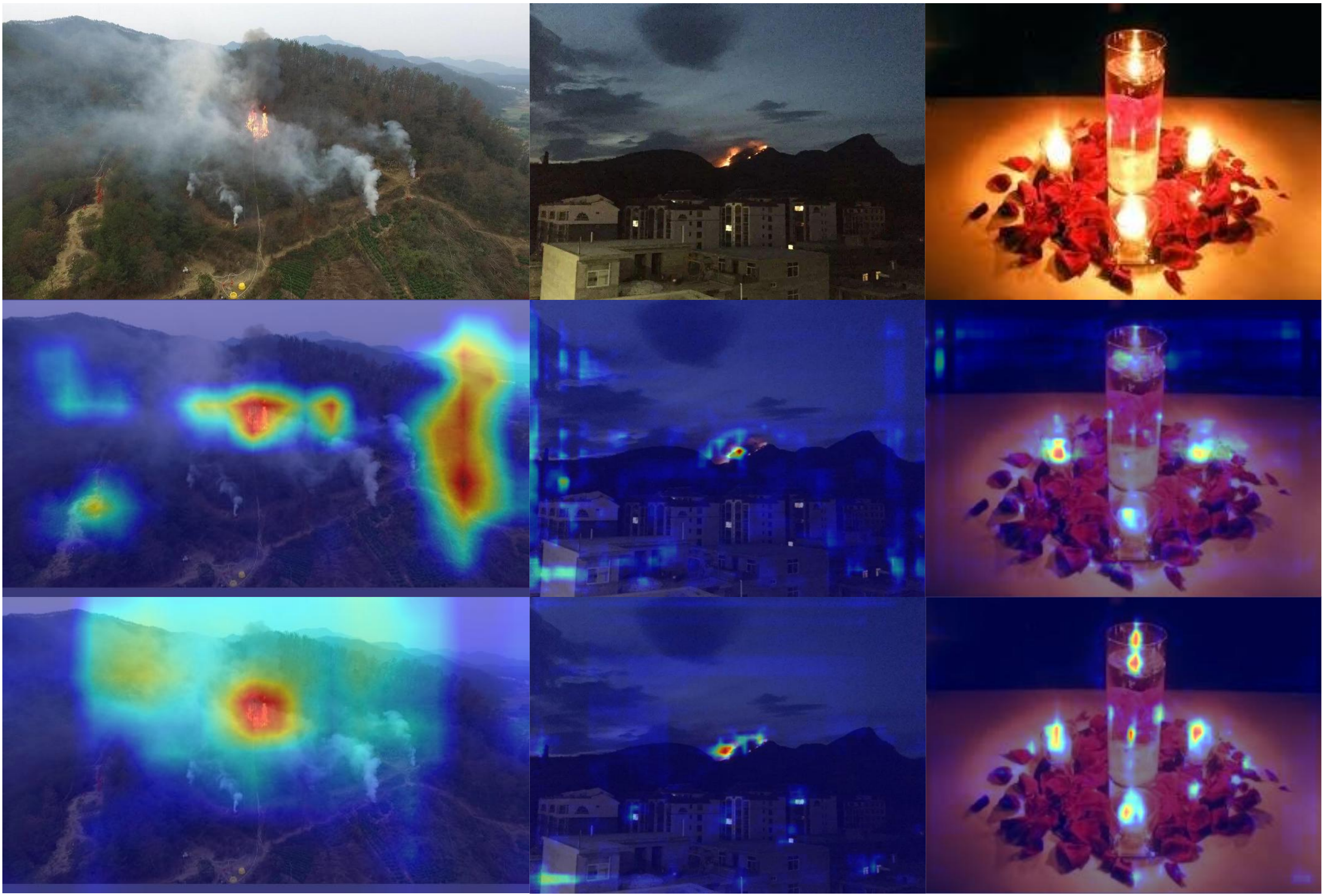}
    \caption{Heat map visualisation results, the first row is the original image, the second row is the YOLOv8n visualisation results and the third row is the YOLO-FireAD visualisation results}
    \label{fig8}
\end{figure}

Figure 8 shows the heat map visualisation results of YOLOv8n and YOLO-FireAD in the flame recognition task. From the figure, it can be seen that although YOLOv8n is able to identify most of the flames, there are some false detections and the identification of the flames is not obvious enough, and there are leakage detections. In contrast, YOLO-FireAD performs much better, accurately identifying the location of flames and more clearly identifying small flames.

\begin{table}[ht]
\centering
\caption{Test results of generalisability of all models verified at Fire dataset}
\label{tab2}
\begin{tabular}{@{}lcccc@{}}
\toprule
\textbf{Model} & \textbf{P} & \textbf{R} & \textbf{mAP50} & \textbf{mAP50-95} \\
\midrule
YOLOv8n & 37.0 & 28.8 & 25.6 & 7.83 \\
YOLOv9t & 29.8 & 28.5 & 22.4 & 6.73 \\
YOLOv10n & 34.4 & 29.4 & 22.7 & 6.99 \\
YOLO11n & 40.7 & 31.6 & 27.2 & 7.85 \\
YOLOv12n & 39.7 & 27.1 & 24.5 & 8.0 \\
YOLOv8n+SE & 43.4 & 22.5 & 22.8 & 7.24 \\
YOLOv8n+HAM & 35.5 & 28.6 & 22.5 & 6.75 \\
YOLOv8n+ASF & 39.0 & 28.3 & 25.9 & 7.48 \\
YOLOv8n+BifPN & 38.7 & 31.5 & 25.8 & 7.8 \\
YOLOv8n+WIoUv3 & 33.1 & 28.0 & 21.4 & 5.87 \\
YOLOv8n+MDPIoU & 32.1 & 28.9 & 21.6 & 7.23 \\
YOLOv8n+PIoUv2 & 35.0 & 29.0 & 23.6 & 7.22 \\
\textbf{Ours} & \textbf{35.9} & \textbf{25.7} & \textbf{25.0} & \textbf{8.12} \\
\bottomrule
\end{tabular}
\end{table}

Furthermore, we verify the generalization of the model on the Fire dataset, and the results are shown in Table 2. It can be seen that YOLO-FireAD also performs well. The mAP50-95 has the best performance among all the comparison models, reaching 8.12\%, which is higher than YOLOv9t, YOLOv10n, YOLOv8n+HAM, YOLOv8n+WIoUv3, 1.39\%, 1.13\%, 1.37\%, and 2.25\%, respectively.

\subsection{Ablation study analysis}

\begin{table}[ht]
\centering
\caption{Test results of YOLO-FireAD ablation experiment on fire\_detection dataset.}
\label{tab3}
\begin{tabular}{lccccccccc}
\toprule
\textbf{AIR} & \textbf{DPDF} & \textbf{P} & \textbf{R} & \textbf{mAP50} & \textbf{mAP50-95} & \textbf{Params} & \textbf{GFLOPs} & \textbf{Size} \\
\midrule
 &  & 64.6 & 62.4 & 64.2 & 32.8 & 3.01 & 8.1 & 6.1 \\
\checkmark &  & 68.0 & 56.3 & 63.0 & 32.6 & 1.84 & 5.4 & 3.9 \\
 & \checkmark & 69.2 & 63.8 & 65.9 & 34.5 & 2.52 & 6.9 & 5.2 \\
\checkmark & \checkmark & 75.3 & 53.3 & 64.9 & 34.6 & 1.45 & 4.6 & 3.3 \\
\bottomrule
\end{tabular}
\end{table}

In order to further demonstrate the effectiveness of the two modules proposed in YOLO-FireAD, we conducted ablation experiments on the fire\_detection dataset with the AIR, DPDF modules using YOLOv8n as the baseline algorithm, and the results of the experiments are shown in Table 3. Although the mAP50-95 of YOLO-FireAD decreased slightly after only introducing the AIR module, P increased from 64.6\% to 68.0\%, an increase of nearly 3.4\%. At the same time, Params, FLOPs and Model Size were significantly reduced, especially Params was reduced from 3.01 to 1.84. That's a reduction of about 39 percent. This proves the role that AIR plays in lightweight.

When only the DPDF module is introduced into YOLO-FireAD, compared with YOLOv8n, the P,R,mAP50,mAP50-95 of the model are all improved, among which the improvement of P is the most obvious, from 64.6\% to 69.2\%, and the reductions of Params, FLOPs, and Model Size are also achieved. This proves the usefulness of the DPDF module in improving the model recognition accuracy.

When the two are combined, the overall model performance is optimal, with P further improved to 75.3\%, which is 10.7\% higher compared to YOLOv8n, while mAP50-95 is improved from 32.8\% to 34.6\%. Params, FLOPs, and Model Size are all reduced to 1.45M, 4.6G, and 6.1MB respectively, which can be inferred that both AIR and DPDF play an indispensable role in YOLO-FireAD. The combination of the two makes YOLO-FireAD able to significantly increase the detection rate while guaranteeing the accuracy of the model through the enhancement of key features and the optimal fusion of multi-granularity features.

\subsection{Limitation analysis}

Although the YOLO-FireAD model performs well in flame detection tasks, it still has some limitations. For example, although the AIR module can effectively enhance the response in the flame core region, it may misdetect fire-like interferences in the background (such as light reflections, glare areas, etc.) under complex backgrounds or extreme lighting conditions, leading to a decrease in detection accuracy. In addition, although the DPDF module performs well in multi-granularity feature fusion, when dealing with dynamic scenes (e.g., fast moving flames or fast spreading smoke), inaccurate feature fusion may occur, which affects the detection performance of dynamic flames. Meanwhile, due to the relatively small number of samples in the training data for some specific scenes (such as small target flames or smoke occlusion scenes), the model's detection capability in these scenes may be somewhat limited.

\section{Conclusion}

This study presents YOLO-FireAD, an efficient fire detection framework that addresses critical challenges in real-world deployment through two novel modules: the Attention-guided Inverted Residual Block (AIR) and Dual Pool Downscale Fusion Block (DPDF). AIR integrates a lightweight channel-spatial hybrid attention mechanism into inverted residual learning, enabling adaptive feature recalibration while suppressing fire-like interference. DPDF mitigates feature degradation during downsampling by fusing max-pooling’s edge-preserving capability and average-pooling’s contextual consistency. Extensive experiments across two benchmarks demonstrate the framework’s superiority, outperforming YOLOv8 by 1.8\% mAP with 51.8\% fewer parameters. 


\section{Acknowledgement}

We would like to acknowledge that in the previous version of this work (ICIC 2025), Zhen Tian was inadvertently omitted from the author list due to an oversight during submission. We sincerely apologize for this omission. In this updated version, the author list has been corrected to include Zhen Tian, whose contributions were essential to the completion of this work.

\end{document}